\pdfoutput=1

\documentclass[11pt]{article}

\usepackage{acl}

\usepackage{url}

\usepackage{times}
\usepackage{latexsym}

\usepackage[T1]{fontenc}
\usepackage[utf8]{inputenc}
\usepackage{comment}

\usepackage{microtype}

\usepackage{amsmath, amssymb}
\usepackage{xcolor}
\usepackage{todonotes}
\usepackage{booktabs}
\usepackage{multirow}
\usepackage{graphicx}
\usepackage{float}
\usepackage{subcaption}

\newcommand{\luis}[1]{\textcolor{blue}{#1}}
\newcommand{\devansh}[1]{\textcolor{red}{#1}}
\newcommand{\todoin}[1]{\todo[inline,size=\small,color=red!10]{\textbf{all: }#1}\xspace}

\DeclareMathOperator*{\argmax}{arg\,max}

\title{Distilling Hypernymy Relations from Language Models: On the Effectiveness of Zero-Shot Taxonomy Induction}

\author{\textbf{Devansh Jain$^{\spadesuit}$, Luis Espinosa Anke$^{\diamondsuit}$}\\
$^{\spadesuit}$ Department of Computer Science and Information Systems, BITS Pilani, India \\
  $^{\diamondsuit}$ Cardiff NLP, School of Computer Science and Informatics, Cardiff University, UK  \\
  $^{\spadesuit}$ \texttt{f20180798@pilani.bits-pilani.ac.in} ,
   $^{\diamondsuit}$ \texttt{espinosa-ankel@cardiff.ac.uk} 
 }  

\begin{document}
\maketitle
\begin{abstract}
In this paper, we analyze zero-shot taxonomy learning methods which are based on distilling knowledge from language models via prompting and sentence scoring. We show that, despite their simplicity, these methods outperform some supervised strategies and are competitive with the current state-of-the-art under adequate conditions. We also show that statistical and linguistic properties of prompts dictate downstream performance.
\end{abstract}

\section{Introduction}
\label{sec:intro}

Taxonomy learning (TL) is the task of arranging domain terminologies into hierarchical structures where terms are nodes  and edges denote \textit{is-a} (hypernymic) relationships \cite{Hwangetal2012}. Domain-specific concept generalization is at the core of human cognition \cite{Yuetal2015}, and a key enabler in NLP tasks where inference and reasoning are important, e.g.: semantic similarity \cite{Pilehvaretal:2013, yu2014improving}, WSD \cite{Agirreetal2014} and, more recently, QA \cite{joshi2020contextualized} and NLI \cite{chen2020mining}.

Earlier approaches to taxonomy learning focused on mining lexico-syntactic patterns from candidate $($hyponym, hypernym$)$ pairs \cite{Hearst1992,Snowetal2004,KozarevaandHovy2010,BoellaandDiCaro2013,Espinosa-Ankeetal2016AAAI}, clustering \cite{YangandCallan2009}, graph-based methods \cite{FountainandLapata2012,Velardietal2013} or word embeddings \cite{Fuetal2014,Yuetal2015}. These methods, which largely rely on hand-crafted features, are still relevant today, and complement modern approaches exploiting language models (LMs), either via sequence classification \cite{Chen2021ConstructingTF}, or combining contextual, distributed, and lexico-syntactic features \cite{Yu2020STEAMST}. In parallel, several works have recently focused on using LMs as zero-shot tools for solving NLP tasks, e.g., commonsense, relational and analogical reasoning \cite{Petroni2019LanguageMA,bouraoui2020inducing, ushio-etal-2021-bert-is,paranjape2021prompting}, multiword expression (MWE) identification \cite{anke2021evaluating,garcia2021probing}, QA \cite{shwartz2020unsupervised, banerjee2020self}, domain labeling \cite{sainz2021ask2transformers}, or lexical substitution and simplification \cite{zhou2019bert}. Moreover, by tuning and manipulating natural language queries (often referred to as \textit{prompts}),
impressive results have been recently obtained on tasks such as semantic textual similarity, entailment, or relation classification \cite{shin2020eliciting,qin2021learning}.

In this paper, we evaluate LMs on TL benchmarks using prompt-based and sentence-scoring techniques, and find not only that they are competitive with common approaches proposed in the literature (which are typically supervised and/or reliant on external resources), but that they achieve SoTa results in certain domains.

\section{Methodology}
\label{sec:methodology}
We follow \citet{ushio-etal-2021-bert-is} and define a prompt generation function $\tau_p(t_1, t_2)$ which maps a pair of terms and a prompt type $p$ to a single sentence. For instance,
\begin{equation*}
    \begin{aligned}
        \tau_{kind}\text{(``physics'', ``science'')} = \\
        \text{``physics is a kind of science''}
    \end{aligned}
\end{equation*}

\noindent Then, given a terminology $\mathcal{T}$, the goal is to, given an input term $t \in \mathcal{T}$, retrieve its top $k$ most likely hypernyms, (in our experiments, $k \in \{1, 3, 5\}$), using either masked language model (MLM) prompting ($\S$\ref{subsec:restrict}), or sentence-scoring ($\S$\ref{subsec:scorer}).

\subsection{MLM Prompting}
\label{subsec:restrict}

\paragraph{RestrictMLM} \citet{Petroni2019LanguageMA} introduced a ``fill-in-the-blanks'' approach based on cloze statements (or \textit{prompts}) to extract relational knowledge from pretrained LMs. The intuition being that an LM can be considered to ``know'' a fact (in the form of a $<$\emph{subject, relation, object}$>$ triple) such as $<$\emph{Madrid, capital-of, Spain}$>$ if it can successfully predict the correct words when queried with prompts such as ``Madrid is the capital of {\texttt{[MASK]}}''. 
We extend this formulation to define a hypernym retrieval function $f_{R}(\cdot)$ as follows:
\begin{equation}
    \begin{aligned}
        f_{R}(p, t, \textbf{T}) = P({\small{\texttt{[MASK]}}} | \tau_p(t, {\small{\texttt{[MASK]}}})) * \textbf{T}
    \end{aligned}
\end{equation}
where $p$ is a prompt type, and $\textbf{T}$ is a one-hot encoding of the terms $\mathcal{T}$ in the LM's vocabulary. We follow previous works \citep{Petroni2019LanguageMA, kassner-etal-2021-multilingual} and restrict the output probability distribution since this task requires the construction of a lexical taxonomy starting from a fixed vocabulary.

\paragraph{PromptMLM} For completeness, we also report results for an unrestricted variant of \textit{RestrictMLM}, where the LM's entire vocabulary is considered.

\subsection{LMScorer}
\label{subsec:scorer}
Factual (and true) information such as ``Trout is a type of fish'' should be scored higher by a LM than fictitious information such as ``Trout is a type of mammal''. The method for scoring a sentence depends on the type of LM used.

\paragraph{Causal Language Models} Given a sentence $\mathbf{W}$, causal LMs ($\mathcal{C}$) predict token $w_i$ using only past tokens $\mathbf{W}_{<i}$. Thus, a likelihood score can be estimated for each token $w_i$ from the LM's next token prediction. The corresponding scores are then aggregated to yield a score for the sentence $\mathbf{W}$.
\begin{equation}
    s_{\mathcal{C}}(\mathbf{W}) = \exp \left(\sum_{i=1}^{|\mathbf{W}|} log P_{\mathcal{C}}(w_i | \mathbf{W}_{<i})\right)
\end{equation}

\paragraph{Masked Language Models}
Given a sentence $\mathbf{W}$, masked LMs ($\mathcal{M}$) replace $w_i$ by \texttt{[MASK]} and predict it using past and future tokens. Thus, a pseudo-likelihood score can be computed for each token $w_i$ by iteratively masking it and using the LM's masked token prediction \cite{wang-cho-2019-bert, Salazar2020MaskedLM}. The corresponding scores are then aggregated to yield a score for the sentence $\mathbf{W}$.

\begin{equation}
    s_{\mathcal{M}}(\mathbf{W}) = \exp \left(\sum_{i=1}^{|\mathbf{W}|} log P_{\mathcal{M}}(w_i | \mathbf{W}_{\setminus i})\right)
\end{equation}

\noindent Given the above, we can cast TL as a sentence-scoring problem by evaluating the natural fluency of hypernymy-eliciting sentences. Specifically, for each term $t$, we score the sentences generated using $\tau_p(\cdot)$ with every other term $t'$ in the terminology. We then select the term-pair with the highest sentence score and assume that the corresponding term $t'$ is a hypernym of $t$. Formally, we define a hypernym selection function $f_S(\cdot)$ as follows:
 \begin{equation}
    f_{S}(p, t, \mathcal{T}) = \argmax_{t' \in \mathcal{T} \setminus t} [s(\tau_p(t, t'))]
\end{equation}
where $s$ refers to the scoring function determined by the LM used.

\section{Experimental setup}
\label{sec:experiments}

This section covers the datasets and prompts we use in our experiments\footnote{We use PyTorch and the {\small{\texttt{transformers}}} library \cite{wolf-etal-2020-transformers}, as well as {\small{\texttt{mlm-scoring}}} \cite{Salazar2020MaskedLM} (\url{https://github.com/awslabs/mlm-scoring}).}, as well as the different LMs we consider. Concerning evaluation metrics, we report standard precision ($P$), recall ($R$) and $F$-score at the \textit{edge level} \cite{task13semeval2016}.

\paragraph{Dataset Details} We evaluate our proposed approaches on datasets belonging to two TL SemEval tasks (\textit{TExEval-1}, \citet{task17semeval2015} and \textit{TExEval-2}, \citet{task13semeval2016}). Following recent literature, we consider the \textit{equipment} taxonomy from \textit{TExEval-1} and the English-language \textit{environment}, \textit{science} and \textit{food} taxonomies from \textit{TExEval-2}. For the \textit{science} taxonomy, our results are based on an \textit{average of the 3 subsets}, which is in line with previous work. Since these datasets do not come with training data, they are well suited for unsupervised approaches.

\begin{center}
\begin{table}[h]
\centering
\small
\begin{tabular}{@{}l|l|r|r@{}}
\toprule
Domain & Source & \multicolumn{1}{c|}{$V$} & \multicolumn{1}{c}{$E$} \\ \midrule
\textit{environment} & Eurovoc & 261 & 261 \\ \midrule
\multirow{3}{*}{\textit{science}} & Combined & 453 & 465 \\
 & Eurovoc & 125 & 124 \\
 & WordNet & 429 & 452 \\ \midrule
\textit{food} & Combined & 1556 & 1587 \\ \midrule
\textit{equipment} & Combined & 612 & 615 \\ \bottomrule
\end{tabular}
\caption{Taxonomies statistics. Vertices ($V$) and Edges ($E$) are often used as structural measures.}
\label{tab:dataset}
\end{table}
\end{center}

\paragraph{Prompts} We use the following prompts:
\begin{itemize}
    \item \textit{gen.}: [$t_2$] is more general than [$t_1$].
    \vspace{-0.8ex} \item \textit{spec.}: [$t_1$] is more specific than [$t_2$].
    \vspace{-0.8ex} \item \textit{type}: [$t_1$] is a type of [$t_2$].
\end{itemize}
$gen.$ and $spec.$ prompts are hand-crafted templates to encode, in a general way, the hypernymy relationship. The choice of the $type$ prompt, however, comes from a set of experiments involving all \textit{LPAQA} \cite{Jiang2020HowCW} prompts under the ``\textit{is  a  subclass  of}'' category. We do not consider automatic prompt generation techniques \cite{shin2020eliciting} due to the absence of training data.
Note that for each prompt, we replace $t_1$ with the input term so that the task is always to predict its hypernym.

\paragraph{Language Models} We interrogate BERT \cite{Devlin2019BERTPO} and RoBERTa \cite{Liu2019RoBERTaAR} among masked LMs, and GPT2 \cite{radford2019language} among causal LMs. For each LM, we consider two variants corresponding to approximately 117M parameters and 345M parameters.

\section{Results}
\label{sec:results}

Table~\ref{tab:sci-ev} shows the results on \textit{TExEval-2}'s \textit{science} and \textit{environment}. We compare with the current state of the art (\textit{Graph2Taxo}) \cite{Shang2020TaxonomyCO}, as well as with other strong baselines such as \textit{TaxoRL} \cite{Mao2018EndtoEndRL} and \textit{TAXI} \cite{Panchenko2016TAXIAS}, the highest ranked system in \textit{TExEval-2}. We also compare with \textit{CTP} \cite{Chen2021ConstructingTF} to illustrate the advantages of zero-shot methods vs finetuning. For the \textit{environment} domain, we find that \textit{RestrictMLM} performs similar to \textit{CTP} and  \textit{LMScorer} outperforms it. Moreover, all 3 proposed approaches fail to outperform the other baselines. However, in \textit{science}, all 3 of our approaches outperform \textit{CTP}, while our best model (\textit{RestrictMLM}) outperforms \textit{TAXI} and is competitive with \textit{TaxoRL} (ours has higher precison, but lower recall). Note that compared to our zero-shot approaches, these methods are either supervised, expensive to train or take advantage of external taxonomical resources such as WordNet, or lexico-syntactic patterns mined from the web using different hand-crafted heuristics.

\begin{table}[h]
\resizebox{\columnwidth}{!}{
\begin{tabular}{@{}l|rrr|rrr@{}}
\toprule
                              & \multicolumn{3}{c|}{\textit{environment}}                                       & \multicolumn{3}{c}{\textit{science}}                                      \\ \midrule
Model                                & \multicolumn{1}{c}{P} & \multicolumn{1}{c}{R} & \multicolumn{1}{c|}{F} & \multicolumn{1}{c}{P} & \multicolumn{1}{c}{R} & \multicolumn{1}{c}{F} \\ \midrule
\textit{TAXI}       & 33.8                  & 26.8                  & 29.9                   & 35.2                  & 35.3                  & 35.2                  \\
\textit{TaxoRL}     & 32.3                  & 32.3                  & 32.3                   & 37.9                  & 37.9                  & 37.9                  \\
\textit{Graph2Taxo} & 89.0                  & 24.0                  & \textbf{37.0}                   & 84.0                  & 30.0                  & \textbf{44.0}                  \\
\textit{CTP}        & 23.1                  & 23.0                  & 23.0                   & 29.4                  & 28.8                  & 29.1                  \\ \midrule
\textit{PromptMLM}                        & 19.2                  & 19.2                  & 19.2                   & 34.4                  & 32.0                  & 33.1                  \\
\textit{RestrictMLM}                        & 23.0                  & 23.0                  & 23.0                   & 39.3                  & 36.7                  & 37.9                  \\
\textit{LMScorer}                         & 26.4                  & 26.4                  & 26.4                   & 33.1                  & 30.7                  & 31.8                  \\ \bottomrule
\end{tabular}
}
\caption{Comparison of our best performing methods with previous work (\textit{environment} and \textit{science}).}
\label{tab:sci-ev}
\end{table}

We also show results for \textit{TExEval-1}'s \textit{equipment} and \textit{TExEval-2}'s \textit{food} (Table~\ref{tab:food-eq}). Both datasets are considerably larger than \textit{environment} and \textit{science}. We compare with the corresponding highest ranked system, namely \textit{TAXI} for \textit{food}, and \textit{INRIASAC} \cite{grefenstette2015inriasac} for \textit{equipment}. For both domains, all 3 of our approaches outperform the corresponding \textit{TExEval} best-performing systems. This suggests that zero-shot TL with LMs is robust, easily scalable and feasible on large taxonomies.

Finally, a clear bottleneck for prompt-based methods is that only single-token terms can be predicted (using a single {\small{\texttt{[MASK]}}} token), making this approach a lower bound for TL.

\begin{table}[h]
\resizebox{\columnwidth}{!}{
\begin{tabular}{@{}l|rrr|rrr@{}}
\toprule
                           & \multicolumn{3}{c|}{\textit{food}}                                           & \multicolumn{3}{c}{\textit{equipment}}                                   \\ \midrule
Model                             & \multicolumn{1}{c}{P}    & \multicolumn{1}{c}{R} & \multicolumn{1}{c|}{F} & \multicolumn{1}{c}{P} & \multicolumn{1}{c}{R} & \multicolumn{1}{c}{F} \\ \midrule
\textit{TExEval} & \multicolumn{1}{l}{13.2} & 25.1                  & 17.3                   & 51.8                  & 18.8                  & 27.6                  \\ \midrule
\textit{PromptMLM}                     & 23.2                     & 22.6                  & 22.9                   & 29.4                  & 29.3                  & 29.4                  \\
\textit{RestrictMLM}                     & 25.2                     & 24.6                  & \textbf{24.9}                   & 38.4                  & 38.2                  & \textbf{38.3}                  \\
\textit{LMScorer}                    & 25.2                     & 24.6                  & \textbf{24.9}                   & 37.7                  & 37.6                  & 37.6                  \\ \bottomrule
\end{tabular}
}
\caption{Comparison of our best configurations with the best TExEval systems on \textit{food} and \textit{equipment}.}
\label{tab:food-eq}
\end{table}

\begin{table*}[!th]
\resizebox{\textwidth}{!}{
\begin{tabular}{@{}l|l|c|rrr|c|rrr|c|rrr|crrr@{}}
\toprule
 &  & \multicolumn{4}{c|}{\textit{environment}} & \multicolumn{4}{c|}{\textit{science}} & \multicolumn{4}{c|}{\textit{food}} & \multicolumn{4}{c}{\textit{equipment}} \\ \midrule
Method & LM & $(p, k)$ & \multicolumn{1}{c}{P} & \multicolumn{1}{c}{R} & \multicolumn{1}{c|}{F} & $(p, k)$ & \multicolumn{1}{c}{P} & \multicolumn{1}{c}{R} & \multicolumn{1}{c|}{F} & $(p, k)$ & \multicolumn{1}{c}{P} & \multicolumn{1}{c}{R} & \multicolumn{1}{c|}{F} & \multicolumn{1}{c|}{$(p, k)$} & \multicolumn{1}{c}{P} & \multicolumn{1}{c}{R} & \multicolumn{1}{c}{F} \\ \midrule
\multirow{4}{*}{\textit{PromptMLM}} & BERT-Base & $(t, 1)$ & 18.8 & 18.8 & 18.8 & $(t, 1)$ & 30.2 & 28.1 & 29.1 & $(t, 1)$ & 20.9 & 20.4 & 20.6 & \multicolumn{1}{c|}{$(t, 1)$} & 29.4 & 29.3 & 29.4 \\
 & BERT-Large & $(t, 1)$ & 19.2 & 19.2 & 19.2 & $(t, 1)$ & 34.4 & 32.0 & 33.1 & $(t, 1)$ & 23.2 & 22.6 & 22.9 & \multicolumn{1}{c|}{$(t, 1)$} & 28.4 & 28.3 & 28.4 \\
 & RoBERTa-Base & $(t, 1)$ & 18.0 & 18.0 & 18.0 & $(t, 1)$ & 24.5 & 23.0 & 23.7 & $(t, 1)$ & 18.5 & 18.0 & 18.2 & \multicolumn{1}{c|}{$(t, 1)$} & 26.3 & 26.2 & 26.3 \\
 & RoBERTa-Large & $(t, 1)$ & 18.0 & 18.0 & 18.0 & $(t, 1)$ & 28.1 & 26.2 & 27.1 & $(t, 1)$ & 20.3 & 19.8 & 20.0 & \multicolumn{1}{c|}{$(t, 1)$} & 28.4 & 28.3 & 28.4 \\
\midrule
\multirow{4}{*}{\textit{RestrictMLM}} & BERT-Base & $(t, 1)$ & 23.0 & 23.0 & 23.0 & $(t, 1)$ & 35.8 & 33.5 & 34.6 & $(t, 1)$ & 22.8 & 22.2 & 22.5 & \multicolumn{1}{c|}{$(t, 1)$} & 38.4 & 38.2 & 38.3 \\
 & BERT-Large & $(t, 1)$ & 21.8 & 21.8 & 21.8 & $(t, 1)$ & 39.3 & 36.7 & \textbf{37.9} & $(t, 1)$ & 25.2 & 24.6 & \textbf{24.9} & \multicolumn{1}{c|}{$(t, 1)$} & 37.9 & 37.7 & \textbf{37.8} \\
 & RoBERTa-Base & $(t, 1)$ & 5.4 & 5.4 & 5.4 & $(t, 1)$ & 11.0 & 10.6 & 10.8 & $(t, 1)$ & 9.3 & 9.1 & 9.2 & \multicolumn{1}{c|}{$(t, 1)$} & 0.0 & 0.0 & 0.0 \\
 & RoBERTa-Large & $(t, 1)$ & 8.4 & 8.4 & 8.4 & $(t, 1)$ & 12.3 & 11.8 & 12.0 & $(t, 1)$ & 10.7 & 10.5 & 10.6 & \multicolumn{1}{c|}{$(t, 1)$} & 0.0 & 0.0 & 0.0 \\
\midrule
\multirow{6}{*}{\textit{LMScorer}} & BERT-Base & $(t, 1)$ & 20.3 & 20.3 & 20.3 & $(t, 1)$ & 15.2 & 14.4 & 14.8 & $(t, 3)$ & 6.8 & 19.7 & 10.1 & \multicolumn{1}{c|}{$(t, 3)$} & 7.5 & 22.4 & 11.2 \\
 & BERT-Large & $(t, 3)$ & 13.7 & 41.0 & 20.5 & $(t, 1)$ & 13.0 & 12.4 & 12.6 & $(t, 1)$ & 13.9 & 13.6 & 13.7 & \multicolumn{1}{c|}{$(t, 1)$} & 15.2 & 15.1 & 15.1 \\
 & RoBERTa-Base & $(g, 3)$ & 7.7 & 23.0 & 11.5 & $(t, 3)$ & 5.5 & 15.7 & 8.1 & $(t, 3)$ & 2.5 & 7.2 & 3.7 & \multicolumn{1}{c|}{$(t, 5)$} & 4.2 & 21.0 & 7.0 \\
 & RoBERTa-Large & $(t, 3)$ & 11.1 & 33.3 & 16.7 & $(t, 1)$ & 13.6 & 12.8 & 13.2 & $(t, 3)$ & 3.6 & 10.6 & 5.4 & \multicolumn{1}{c|}{$(t, 3)$} & 9.2 & 27.5 & 13.8 \\
 & GPT-2 Base & $(t, 1)$ & 24.9 & 24.9 & 24.9 & $(t, 1)$ & 29.3 & 27.4 & 28.3 & $(t, 1)$ & 21.0 & 20.5 & 20.7 & \multicolumn{1}{c|}{$(t, 1)$} & 36.8 & 36.6 & 36.7 \\
 & GPT-2 Medium & $(t, 1)$ & 26.4 & 26.4 & \textbf{26.4} & $(t, 1)$ & 33.1 & 30.7 & 31.8 & $(t, 1)$ & 25.2 & 24.6 & \textbf{24.9} & \multicolumn{1}{c|}{$(t, 1)$} & 37.7 & 37.6 & 37.7 \\ \bottomrule
\end{tabular}
}
\caption{Comparison of best configuration for each LM and proposed approach. $(p, k)$ refers to the prompt and top-$k$ combination that gives the best results for that setting, where $p$ = $g$ for \textit{gen.}, $s$ for \textit{spec.} and $t$ for \textit{type} prompt.}
\label{tab:lm-compare}
\end{table*}

\section{Analysis}
\label{sec:analysis}

In this section, we provide an in-depth analysis of our approaches, including comparison of LMs and statistical and semantic properties of prompts.

\paragraph{LM Comparison} Table \ref{tab:lm-compare} compares the best configuration for each LM. We can immediately see that a conservative approach (i.e., $k=1$ with the \textit{type} prompt) almost always yields the best $F$-score. Another important conclusion is that, among MLMs, BERT-Large performs best across the board, with BERT generally outperforming RoBERTa, a finding in line with previous works \cite{shin2020eliciting}. Concerning causal LMs, GPT-2 Medium outperforms its smaller counterpart as well as both MLMs for sentence-scoring.

\paragraph{Sensitivity to Prompts} There is interested in understanding the model's sensitivity to prompts and whether frequency can explain downstream performance in lexical semantics tasks \cite{chiang2020understanding}. In the context of prompt vs. performance correlation, we find that prompt-based downstream performance on TL can be attributed to: (1) syntactic completeness and (2) semantic correctness. For (1), we find that prompts that are syntactically more complete (e.g., ``\textit{[X] is a type of [Y]}'' vs ``\textit{[X] is a type [Y]}'', the difference being the prepositional phrase) perform better. For (2), we find that prompts that unambiguously encode hypernymy are also better (i.e., the \textit{type} prompt, as opposed to other noise-inducing templates such as ``\textit{is a}'' or ``\textit{is kind of}''). Finally, out of the cleanest prompts, the most frequent in pretraining corpora are the most competitive. Table \ref{tab:freqs} confirms the intuition that the \textit{type} prompt is not only unambiguous, but also highly frequent when compared to similar (noise-free and syntactically complete) prompts.

\begin{center}
\begin{table}[!h]
\small
\centering
\begin{tabular}{@{}l|r||r@{}}
\toprule
Prompt & \multicolumn{1}{c||}{\textit{avg} F} & \multicolumn{1}{c}{Frequency} \\ \midrule
is a type of & 25.5 & 14,503 \\
is the type of & 24.2 & 809 \\
is a kind of & 23.6 & 2,934 \\
is a form of & 22.1 & 9,518 \\
is one form of & 17.9 & 124 \\
is a & 7.4 & 9,328,426 \\
is a type & 1.0 & 15,085 \\ \bottomrule
\end{tabular}
\caption{Domain-wise average $F$-score of LPAQA prompts and their frequency in BERT's pretraining corpora.}
\label{tab:freqs}
\end{table}
\end{center}

\paragraph{Single-Token vs Multi-Token Hypernyms} 
Table~\ref{tab:single-stats} compares \textit{F-score} on original terminology vs filtered terminology, where filtered terminology contains only the terms that have single-token hypernyms. The results show that \% Increase in \textit{F-score} is inversely proportional to the \% Retained. This can be explained by the fact that smaller \% of terms retained implies higher \% of multi-token hypernyms in the original dataset that cannot be predicted using prompting. Thus, the increase in \textit{F-score} by removing such hypernyms should increase as the \% Retained decreases.

\begin{table}[h]
\resizebox{\columnwidth}{!}{
\begin{tabular}{@{}l|r|r|r@{}}
\toprule
Domain               & \multicolumn{1}{l|}{Total Terms} & \multicolumn{1}{l|}{\% Retained} & \multicolumn{1}{l}{\% Increase} \\ \midrule
\textit{environment} & 261                              & 29.89                            & 2.32                            \\
\textit{equipment}   & 612                              & 44.77                            & 1.24                            \\
\textit{science}     & 452                              & 53.32                            & 0.90                            \\
\textit{science\_ev} & 125                              & 52.80                            & 0.89                            \\
\textit{food}        & 1555                             & 59.55                            & 0.57                            \\
\textit{science\_wn} & 370                              & 69.73                            & 0.51                            \\ \bottomrule
\end{tabular}
}
\caption{Comparison of \textit{F-score} on original terminology vs filtered terminology. \% Retained refers to the percentage of terms that have single-token hypernyms and are thus retained for the filtered dataset. \% Increase shows the increase in \textit{F-score} on filtered dataset compared to \textit{F-score} on original dataset.}
\label{tab:single-stats}
\end{table}

\section{Conclusion and Future Work}
\label{sec:conclusions}
We have presented a study of different LMs under different settings for zero-shot taxonomy learning. Compared with computationally expensive and highly heuristic methods, our zero-shot alternatives prove remarkably competitive. For the future, we could explore multilingual signals and the integration of traditional word embeddings with contextual representations.

\bibliography{custom}
\bibliographystyle{acl_natbib}

\end{document}